\documentclass[preprint]{elsarticle}

\usepackage{lineno,hyperref,natbib,geometry,fleqn,txfonts}

\usepackage{microtype}
\usepackage[english]{babel}
\usepackage[latin1]{inputenc}
\usepackage{graphicx}
\usepackage{color}
\usepackage{amsfonts}
\usepackage{mathrsfs}
\usepackage{subfig}
\usepackage{rotating}
\usepackage{algorithm}
\usepackage[noend]{algpseudocode}

\makeatletter
\def\BState{\State\hskip-\ALG@thistlm}
\makeatother

\newtheorem{defi}{Definition}[section]
\newtheorem{exem}{Example}[section]
\newtheorem{prop}{Proposition}[section]
\newtheorem{rmk}{Remark}[section]

\newcommand{\PC}[1]{\ensuremath{\left(#1\right)}}

\begin{document}
	
	\begin{frontmatter}
		
		\title{Combining Multiple Algorithms in Classifier Ensembles using Generalized Mixture Functions\footnote{Preprint submited to Neurocomputing}}
		
		\author[ufrn]{Valdigleis S. Costa\corref{cor2}}
		\ead{valdigleis@gmail.com}
		
		\author[ufrn,ufersa]{Antonio Diego S. Farias\corref{cor1}}
		\ead{antonio.diego@ufersa.edu.br}
		
		\author[ufrn]{Benjam\'in Bedregal}
		\ead{bedregal@dimap.ufrn.br}
		
		\author[ufrn]{Regivan H. N. Santiago}
		\ead{regivan@dimap.ufrn.br}
		
		\author[ufrn]{Anne Magaly de P. Canuto}
		\ead{anne@dimap.ufrn.br}

		\cortext[cor1]{Corresponding author}
		
		\cortext[cor2]{Principal corresponding author}
		
		\address[ufrn]{Department of Informatics and Applied Mathematics, Federal University of Rio Grande do Norte (UFRN), Natal - RN, 59.072-970}
		
		\address[ufersa]{Department of Exacts and Naturals Sciences, Federal Rural University of Semi-Arid (UFERSA), Pau dos Ferros - RN, 59.900-000}
		
		\begin{abstract}
			Classifier ensembles are pattern recognition structures composed of a set of classification algorithms (members), organized in a parallel way, and a combination method with the aim of increasing the classification accuracy of a classification system. In this study, we investigate the application of a generalized mixture (GM) functions as a new approach for providing an efficient combination procedure for these systems through the use of dynamic weights in the combination process. Therefore, we present three GM functions to be applied as a combination method. The main advantage of these functions is that they can define dynamic weights at the member outputs, making the combination process more efficient.  In order to evaluate the feasibility of the proposed approach, an empirical analysis is conducted, applying classifier ensembles to 25 different classification data sets. In this analysis, we compare the use of the proposed approaches to ensembles using traditional combination methods as well as the state-of-the-art ensemble methods. Our findings indicated gains in terms of performance when comparing the proposed approaches to the traditional ones as well as comparable results with the state-of-the-art methods. 
			
		\end{abstract}
		
		
	\end{frontmatter}

\section{Introduction}

In machine learning, a classifier ensemble, also known as ensemble systems, ensemble of classifiers or simply ensembles, can be understood as a collaborative decision-making system composed of $N$ members (individual classifiers), in which a strategy is applied to combine the predictions of ensemble members to generate a single prediction as output \cite{chen01}. In other words, a classifier ensemble is a two-layer pattern recognition structure in which the first layer is composed of a set of $N$ individual classifiers and the second layer is composed of a combination module \cite{kiter02}. Essentially, the combination module is responsible for combining the outputs of the individual classifiers and for transforming them into a single output, which is the final output of an ensemble. The use of classifier ensembles in machine learning is not recent and, as stated in \cite{hoet01}, the first reference that uses classifier ensembles dates back to 1963, in \cite{barn01}. Since then, classifier ensembles have been used in different classification problems, for example, recognition of faces \cite{czyz01}, revocable biometrics \cite{pintro01}, among other applications. 
In addition to classification, there are  several other application domains that the combination of multiple input information has been efficiently applied in order to generate a single output, for example: data clustering \cite{cheng01}, support decision-making \cite{silva01}\cite{silva02}\cite{yager01} and images processing \cite{busti01} \cite{farias01}\cite{farias02}. In this paper, we investigate the use of classifier ensembles in the pattern classification context.



When working with classifier ensembles, one important issue to be taken into consideration is related to the selection of an efficient combination method. 
Ideally, this method should be able to exploit the individual strengths of all individual classifiers and,  at the same time, to minimize their drawbacks \cite{kun04}. 
For many years, simple methods as majority vote \cite{ali01}, linear combination and fusion methods \cite{cha01,kiter01,kiter02,wolp01} were the most popular methods since they were simple and provided reasonable performance. However, with the increase of data complexity, classifier ensembles started to require flexible approaches that can adjust their combination methods to the properties of analyzed datasets. Therefore, the use of trained combiners has gained a significant attention of the machine learning community. Nevertheless, these combination methods require additional training time and access to separate subset of examples and, once again, this requirement can become prohibitive in certain application areas.



One way to improve the efficiency of combination methods is through the use of weights that can be used to denote the confidence (influence) of the individual classifiers in classifying an input pattern to a particular class \cite{kun04}. 
Different ways of calculating weights (confidence) of each class for each individual classifier can be used in determining the relative contribution of each classifier within a classifier ensemble and they can be classified as static \cite{Kim01,Kraw01,Kraw02,Liu01, Onan01,Pintro02, Xiao01} or dynamic weighting  \cite{Bashir01,Kraw01,Onan01,Pintro02}. For offering more flexibility and efficiency, in this paper, we will be working with dynamic weight selection (dynamic weighting).


One possible solution to the selection of an efficient combination method is the use of aggregation functions. They are mathematical models that are capable of solving the task of aggregating several sources of information and generating a single output. Among the most common aggregation functions found in the literature, we can cite t-norms, t-corms \cite{klem01}, aggregation functions \cite{dbois01}\cite{pate01}, among others. In a recent study, \cite{farias01,farias02,farias03,farias04,farias05}, Farias \textit{et al.} have investigated a class of functions called generalized mixture (GM) functions. The GM functions are capable of generalize the notions of ordered weighted averaging (OWA) \cite{yager02} and mixture function \cite{belia01}. The main advantage of GM functions is that the weights of its inputs can be dynamically defined for each input. In other words, while the weights of the other weighted functions are assigned statically, without taking into account the testing patterns, the generalized mixture functions can assign weights as a function of the testing patterns \cite{farias04}. In practical applications, the GM functions have presented good results in the tasks of noise treatment and image reduction \cite{farias01, farias02, farias04}. This advantage can be very useful for classifier ensembles, leading to an efficient decision making process.


Therefore, in this paper, we propose a new combination approach for classifier ensembles using the generalized mixture (GM) functions proposed in \cite{farias01}. In other words, we adapt the GM functions to be used as the combination method of an ensemble system. In this sense, we explore the main advantage of the GM functions, dynamic weighting, in the combination process of an ensembles. Then, the weights related to the decision of each individual classifier is defined dynamically, according to the classifier outputs and the relation among the outputs of all classifiers. 
In order to evaluate the feasibility of the proposed approach, we perform an empirical analysis of ensemble performance in 25 different classification data sets, comparing its performance with classifier ensembles using traditional methods (those presented in \cite{kiter01}) as well as the state-of-the-art ensemble methods. In addition, a statistical analysis is also performed to analyze the performance of classifier ensembles, from the statistical point of view. 

This paper is divided into eight sections and it is organized as follows. In Section 2, we describe some recent studies in classifier ensembles. The fundamental notions of the generalized mixture functions are introduced in Section 3, while the basic concepts of classifier ensembles are presented in Section 4. The proposed approach is presented in Section 5. In Section 6, the experimental methodology is presented, while Section 7 presents an analysis of the obtained results of this work. Finally, Section 7 concludes this paper.

\section{Recent Studies in Weighted Combination Methods for in Classifier Ensembles}

As already mentioned, different ways of calculating the weights of each class for each individual classifier can be used in a classifier ensemble \cite{CAO2015,Kim01,Kraw01,Kraw02,Liu01, Onan01,Pintro02, Xiao01}. Although weighted combination methods appear to provide some flexibility, obtaining the optimal weights is not an easy task. Therefore, some optimization techniques have been applied to define the best set of weights, such as in \cite{Bashir01,Kraw01,Onan01,Pintro02}. In \cite{Pintro02}, for instance, a genetic algorithm (GA) was used to define an optimized set of weights that are used along with the output of the individual classifiers to define the final output of the classifier ensembles. However, all the aforementioned studies apply procedures to define static weights. In other words, these methods define a set of weights that are used throughout the testing phase. This static way to define weights can eventually become inefficient for a classifier ensemble, since the accuracy of an individual classifier can change in the testing search space and this change is not capture by static weights.

In a dynamic weighting process, the outputs of all individual classifiers are aggregated and the most competent ones receive the highest weight values. The competence of the classifier outputs are usually based on some local competence measure. There are some studies that apply dynamic weights in combination methods, such as in \cite{CAO2015,JIMENEZ1998,KOLTER2003,Pagano01,Par01,Pol01,Wol01}. However, in most of these studies, the dynamic weighting process relies on a model that has to be built as a neural network \cite{JIMENEZ1998}, a histogram representation \cite{Pagano01}, quadratic programming \cite{Pol02}, fuzzy classifier \cite{Stef01}, among others. These models usually requires an extra processing to be built and become complex structures to be designed. In addition, in \cite{XU2015,KOLTER2003}, adaptive mechanisms are applied. In \cite{KOLTER2003}, for instance, a dynamic weighted majority (DWM) method is proposed, in which it uses a weighted-majority vote of the classifier and dynamically creates and deletes classifiers in response to changes in performance. However, the dynamic weights are defined based on the performance of the classifiers based on previously seen instances and not based on the information of the current instance to be classified.


In addition, some studies proposed a dynamic weighting procedure for a specific domain, such as concept drift \cite{KOLTER2003,Pagano01,SONG2016}, textual and visual content-based anti-phishing \cite{ZANG2011}, among others. In the dynamic weighting technique proposed in \cite{Pagano01}, for instance, each classifier is dynamically weighted based on the similarity between an input pattern and the histogram representation of each concept present in the ensemble. In the mentioned paper, the Hellinger distance between an input and the histogram representation of every previously-learned concept is computed, and the score of every classifier is weighted dynamically according to the resemblance to the underlying concept distribution. According to the authors, the empirical analysis  with synthetic problems indicate that the proposed fusion technique is able to increase system performance when input data streams incorporate abrupt concept changes, yet maintains a level of performance that is comparable to the average fusion rule when the changes are more gradual. 

There are also some studies that are limited to a data-dependent measures. For instance, in \cite{Par01}, the authors used ad-hoc data-dependent measures in the dynamic weight setting procedure and noisy data could compromise the overall performance of the ensemble system.

Unlike the aforementioned studies, in this paper, we present a family of aggregation  functions (GM) that is adapted to be used in a dynamic weighting procedure for classifier ensembles. In other words, these functions define weights for the output of the individual classifiers, in a dynamic way, without having to build a model and using information of the current instance to be classified. These functions are inexpensive and straightforward in system design and setup, leading to accurate and robust classifier ensembles.

\section{Mathematical Framework}

In this section, the mathematical background used in this paper is described, starting with the description of aggregation functions, then going through the description of ordered weighted averaging functions until the definition of generalized mixture functions.

\subsection{Aggregation functions}

Aggregation functions are mathematical tools that have the ability to combine multiple attributes into one single output. These functions are able to transform $n$ attributes belonging to the $[0,1]$ interval into a single attribute also in the $[0,1]$ interval. More precisely,

\begin{defi}\label{defi3.1}
	A {\bf $n$-dimensional aggregation function} or simply {\bf aggregation} is a monotonic\footnote{A function $A:[0,1]^n\longrightarrow [0,1]$ is monotonic if $A(x_1,\cdots,x_n)\le A(y_1,\cdots,y_n)$ whenever $x_i\le y_i$ for all $i=1,\cdots,n$.} function $A:[0,1]^n\longrightarrow[0,1]$ that satisfies the boundary condition: $A(0,\cdots,0)=0$ and $A(1,\cdots,1)=1$.
\end{defi}

The {\it Maximum}, {\it Minimum}, {\it Arithmetic Mean} and {\it Product} functions, described in the following example, are classical examples of aggregations.

\begin{exem}\label{exem3.1}
	The reader can easily prove that the functions below are aggregations ones:
	\begin{enumerate}
		\item[(a)] $max(x_1,\cdots,x_n)=max\{x_1,\cdots,x_n\}$;
		\item[(b)] $min(x_1,\cdots,x_n)=min\{x_1,\cdots,x_n\}$;
		\item[(c)] $arith(x_1,\cdots,x_n)=\frac{\sum\limits_{i=1}^n x_i}{n}$;
		\item[(d)] $prod(x_1,\cdots,x_n)=\prod\limits_{i=1}^nx_i$.
	\end{enumerate}
\end{exem}

More examples of aggregation functions can be found in \cite{belia01}. In addition, applications of aggregation functions can be found in several research fields as, for instance, it can be used to model the connectives of fuzzy logic. Among these connectives, it is important to highlight the t-norms and t-conorms functions \cite{klem01}.

Aggregation functions can satisfy a wide range of properties, not restricted to the monotonicity and boundary conditions. In the definition below, some of these properties are listed.

\begin{defi}
	Let $A:[0,1]^n\longrightarrow[0,1]$ be an aggregation function. We say that $A$:
	\begin{enumerate}
		\item is {\bf idempotent} if, and only if $A(x,\cdots,x)=x$, for all $x\in [0,1]$;
		\item is {\bf homogeneous of order $k$} if, and only if $A(\lambda x_1,\cdots,\lambda x_n)=\lambda^k\cdot A(x_1,\cdots,x_n)$, for any $\lambda \in[0,1]$ and $(x_1,\cdots,x_n)\in [0,1]^n$;
		\item has a {\bf neutral element} if there is an element $e\in[0,1]$ such that $A(e,\cdots,e,x_i,e,\cdots,e)=x_i$, for all $x_i\in[0,1]$ and any coordinate $i\in\{1,\cdots,n\}$;
		\item has a {\bf absorbing element} or {\bf annihilator} if there is an element $a\in[0,1]$ such that for all $i\in\{1,\cdots,n\}$ and $(x_1,\cdots,x_{i-1},a,x_{i+1},\cdots,x_n)\in[0,1]^n$, we have $A(x_1,\cdots,x_{i-1},a,x_{i+1},\cdots,x_n)=a$;
		\item is {\bf symmetric} if, and only if $A(\sigma({\bf x})) = A(x_{\sigma(1)},\cdots,x_{\sigma(n)}) = A(x_1,\cdots,x_n)$, for any permutation $\sigma:\{1,\cdots,n\}\rightarrow \{1,\cdots,n\}$ and any $(x_1,\cdots,x_n)\in[0,1]^n$;
		\item is {\bf shift-invariant} if for all $\lambda\in[-1,1]$, chosen property\footnote{That is, $x_i+\lambda\in[0,1]$, for any $i=1,\cdots,n$, and $A(x_1,\cdots,x_n)+\lambda\in[0,1]$.}, we have $A(x_1+\lambda,\cdots,x_n+\lambda)=A(x_1,\cdots,x_n)+\lambda$;
		\item has a {\bf zero divisor} if there is  $\overrightarrow{x}=(x_1,\cdots,x_n)\in (0,1]^n$ such that $A(\overrightarrow{x})=0$;
		\item has a {\bf one divisor} if there is  $\overrightarrow{x}=(x_1,\cdots,x_n)\in [0,1)^n$ such that $A(\overrightarrow{x})=1$.
	\end{enumerate}
\end{defi}

Of the aggregation functions presented in example \ref{exem3.1}, $min$, $max$ and $arith$ are idempotent, homogeneous of order $1$ (in this case, we say simply homogeneous), symmetric, shift-invariant and do not have zero neither one divisors. In addition, the $min$ function has $1$ as neutral element and $0$ as absorbing element; the $max$ function has $0$ as neutral element and $1$ as absorbing element; the $arith$ function does not have neither neutral elements nor absorbing elements; the $prod$ function has $1$ as neutral element, $0$ as absorbing element, it is symmetric, and does not satisfy any of the other properties.

\begin{rmk}
	Obviously, other functions (which are not aggregations) can also satisfy these properties.
\end{rmk}

\subsection{Ordered Weighted Averaging Functions}

In this section, we present a family of aggregation functions, which was introduced by Yager \cite{yager01} and it is called {\bf Ordered Weighted Averaging} function.

\begin{defi}
	An {\bf Ordered Weighted Averaging} function or {\bf OWA} function is defined by:
	\begin{equation}
	OWA_{\textbf{w}}(x_1, \cdots, x_n) = \displaystyle\sum_{i=1}^{n}w_i x_{(i)}
	\end{equation}
	where $\textbf{w} = (w_1, \cdots w_2)$ is a predetermined vector of weights\footnote{A n-dimensional vector ${\bf w}=(w_1,\cdots,w_n)\in[0,1]^n$ is called of vector of weights if $\displaystyle\sum_{i=1}^{n}w_i = 1$.} and $(x_{(1)}, \cdots, x_{(n)})$ is the decreasing ordering of the input vector $(x_1, \cdots, x_n)$.
\end{defi}

\begin{exem}
	For three coordinates, let ${\bf w}=(0.2,0.45,0.35)$ be the weight vector and $(x_1,x_2,x_3)=(0.7,0,0.3)$ be the input vector. We then have $(x_{(1)},x_{(2)},x_{(3)})=(0.7,0.3,0)$ and OWA is defined as follows.
	
	\begin{equation}
	OWA_{\bf w}(0.7,0,0.3)=0.2\cdot 0.7+0.45\cdot 0.3+0.35\cdot 0=0.275.
	\end{equation}
	
\end{exem}

\begin{exem}
	The functions {\it Minimum}, {\it Maximum}, {\it Arithmetic Mean},  {\it Weighted Averaging Mean} and {\it Median}\footnote{A median formula is given by: $Medi(x_1,\cdots,x_n)=\left\{\begin{array}{ll} x_{(k)}, & \mbox{ if } n=2k-1 \mbox{ and } k\in\mathbb{N}^*\\ \frac{x_{(k)}+x_{(k+1)}}{2}, &\mbox{ if } n=2k \mbox{ and } k\in\mathbb{N}^* \end{array} \right.$, where $\mathbb{N}^*=\{n\in\mathbb{N}:\ n>0\}$ and $x_{(k)}$ is the $k$-th lowest coordinate of $(x_1,\cdots,x_n)$} are example of OWA functions. A more in-depth approach can be found in \cite{farias01,farias02}.
\end{exem}

In the following propositions, we present some properties of OWA functions (proofs of these propositions can be found in \cite{yager01}).

\begin{prop}
	Let ${\bf w}=(w_1,\cdots,w_n)$ be a weight vector. Then, $OWA_{\bf w}$ is idempotent, symmetric, shift-invariant and continuous aggregation function, that does not have one and zero divisors.
\end{prop}

\begin{prop}
	For any  weight vector ${\bf w}=(w_1,\cdots,w_n)$, $OWA_{\bf w}$ is an averaging function. In other words, $min(x_1,\cdots,x_n)\le OWA_{\bf w}(x_1,\cdots,x_n)\le max(x_1,\cdots,x_n)$.
\end{prop}

Although the aggregation functions have proved to be quite effective in performing the task of encoding multiple information into a single one, other functions have also been widely used in this field. One example is the {\bf preaggregation} functions, introduced by Lucca {\it et al.} in \cite{lucca}, that are functions of $[0,1]^n$ to $[0,1]$ that satisfy the boundary condition, but instead of monotonicity, they satisfy the ${\bf r}$-monotonicity \cite{preaggregation}.

\begin{defi}
	If there is a directional vector ${\bf r}=(r_1,\cdots,r_n)$, such that for all $k\geq 0$ and $(x_1,\cdots,x_n)\in[0,1]^n$ with $(x_1+kr_1,\cdots,x_n+kr_n)\in[0,1]^n$ we have $F(x_1,\cdots, x_n)\le F(x_1+kr_1,\cdots,x_n+kr_n)$, then we say that $F$ is {\bf ${\bf r}$-increasing} or that $F$ satisfies the {\bf ${\bf r}$-monotonicity} property. 
\end{defi}

\begin{defi}
	A function $F:[0,1]^n\longrightarrow[0,1]$ is a ${\bf r}$-preaggregation function if it is ${\bf r}$-increasing (for some direction ${\bf r}$) and it satisfies the boundary condition, as Definition \ref{defi3.1}.
\end{defi}

In \cite{Bustince2016,lucca}, examples of preaggregation functions are presented. In the next subsection, we present the generalized mixture functions, that is also an example of a preaggregation function.

\subsection{Generalized Mixture functions}
\label{sec:prop}

As mentioned previously, OWA functions belong to the class of averaging aggregation functions. However, unlike the weighted arithmetic mean function, whose weights are associated with the particular inputs, each OWA weight is associated with the instance magnitude obtained by its input vector. In other words, the importance of each instance is determined by the input vector itself. Generalized Mixture functions (or \textsf{GM} functions), introduced by Pereira \textit{et al.} \cite{pere01}\cite{pere02}, also have the characteristic of designating the importance of an input based on the input vector itself. Nevertheless, unlike in \textsf{OWA}, weights are not fixed and they are defined in a dynamic way in \textsf{GM}. Below, we present the definition of generalized mixture functions.

\begin{defi}
	A finite family of functions $\Gamma = \{f_i:[0,1]^n \rightarrow [0,1]|\ 1 \leq i \leq n\}$, with $\displaystyle\sum_{i=1}^{n}f_i(x_1, \cdots, x_n) = 1$, for all $(x_1, \cdots, x_n) \in [0,1]^n$, is called of {\bf family of weight-functions} (FWF). The {\bf Generalized Mixture} function or {\bf GM} function associated to $\Gamma$, denoted by $GM_\Gamma$, is the function $GM_{\Gamma}:[0,1]^n\longrightarrow[0,1]$ defined by:
	{\small
		\begin{equation}\label{eq:2}
			GM_\Gamma(x_1, \cdots, x_n) = \displaystyle \sum_{i=1}^{n}f_i(x_1, \cdots, x_n)\cdot x_i
		\end{equation}}
\end{defi}

\begin{exem}
	The functions {\it Minimum}, {\it Maximum} and {\it Arithmetic Mean} are examples of $GM_\Gamma$, with FWF for these functions are respectively:
	\begin{enumerate}
		\item[(a)] $f_{(i)}(x_1,\cdots,x_n)=0$, for all $i=1,\cdots,n-1$, and $f_{(n)}(x_1,\cdots,x_n)=1$;
		\item[(b)] $f_{(1)}(x_1,\cdots,x_n)=1$ and $f_{(i)}(x_1,\cdots,x_n)=0$, for all $i=2,\cdots,n$;
		\item[(c)]$f_i(x_1,\cdots,x_n)=\frac{1}{n}$, for all $i=1,\cdots,n$;
		\item[(d)] More generally, any OWA function is also a GM function, and therefore, {\it Median} is also a Generalized Mixture function \cite{farias01}.
	\end{enumerate}
\end{exem}

\begin{rmk}
	The index $(1),(2),\cdots,(i),\cdots,(n)$ of items (a) and (b) from the previous example are obtained by the decreasing ordering $(x_{(1)},\cdots,x_{(n)})$ of vector $(x_1,\cdots,x_n)$.
\end{rmk}

Note that according to item (d) of the previous example, any OWA function is also a GM function. But obviously, the reciprocal is not true. In addition, although GM functions satisfy the boundary condition, they are not always monotonic functions (as shown in the below example). Therefore, a GM function may not be an aggregation one.

\begin{exem}
	Let $f_1(x,y,z)=\left\{\begin{array}{ll}\frac{1}{3},& \mbox{ if } x=y=z=0\\ \frac{x}{x+y+z},& \mbox{ otherwise.} \end{array} \right.$, $f_2(x,y,z)=\left\{\begin{array}{ll}\frac{1}{3},& \mbox{ if } x=y=z=0\\ \frac{y}{x+y+z},& \mbox{ otherwise.} \end{array} \right.$ and also	
	\linebreak$f_3(x,y,z)=\left\{\begin{array}{ll}\frac{1}{3},& \mbox{ if } x=y=z=0\\ \frac{z}{x+y+z},& \mbox{ otherwise.} \end{array} \right.$, then $\Gamma=\{f_1,f_2,f_3\}$ is a weight function with
	$$GM_{\Gamma}(x,y,z)=\left\{\begin{array}{ll}0,& \mbox{ if } x=y=z=0\\ \frac{x^2+y^2+z^2}{x+y+z},& \mbox{ otherwise.} \end{array} \right.$$
	This function satisfies the boundary condition, but it does not satisfy the monotonicity property. Since, $GM(0.5,0.2,0.1)=0.375$ and $GM(0.5,0.22,0.2)=0.368$.
\end{exem}

Farias {\it et al.} \cite{farias01} have proven that the generalized mixture (GM) functions can satisfy several interesting properties. For the purpose of this paper, the most important property is the fact that GM functions satisfy the relation, $min(x_1, \cdots, x_n) \leq GM_\Gamma(x_1, \cdots, x_n) \leq max(x_1, \cdots, x_n)$. In addition,  Farias {\it et al.} provided an special GM function presented in the example 3.6, which made it possible to build a family of GM in \cite{farias02}.

\begin{exem}
	The function $\mathbf{H}:[0,1]^n\longrightarrow [0,1]$ defined by:
	\begin{equation}\label{eq3}
		\mathbf{H}(x_1,\cdots,x_n)=\left\{\begin{array}{ll} x, & \mbox{ if } (x_1,\cdots,x_n)=(x,\cdots,x)\\ \frac{1}{n}\sum\limits_{i=1}^n x_i-\frac{x_i|x_i-Med(x_1,\cdots,x_n)|}{\sum\limits_{j=1}^n|x_j-Med(x_1,\cdots,x_n)|}, & \mbox{ otherwise}\end{array}\right.,
	\end{equation}
	where
	\begin{equation}
		Med(x_1,\cdots,x_n)=\left\{\begin{array}{ll} x_{(k)}, & \mbox{ if } n=2k-1\\ \frac{x_{(k)}+x_{(k+1)}}{2}, & \mbox{ if } n=2k  \end{array}\right.
	\end{equation}
	is a generalized mixture function.
\end{exem}

In the following propositions, we present a series of properties that the $\mathbf{H}$ function satisfies. For simplicity reasons, proofs of these propositions are omitted. However, the reader can check the validity of all listed properties in \cite{farias01}. Additionally, it is possible to prove that $\mathbf{H}$ is a preaggregation.

\begin{prop}
	The function $\mathbf{H}$ is idempotent, homogeneous, shift-invariant, symmetric, does not have neutral and absorbing elements and does not have one and zero divisors.
\end{prop}

Although it has been proven that $\mathbf{H}$ satisfies all these properties described in the previous proposition, the complexity of this function blocked the verification of the monotonicity property. 

\begin{prop}
	$\mathbf{H}$ is a preaggregation function.
\end{prop}

{\it Proof:} $\mathbf{H}$ is $(k,\cdots,k)$-increasing, which is based on the fact that $\mathbf{H}$ is shift-invariant.

\hfill

Besides the $\mathbf{H}$ function, Farias \textit{et al.} established in \cite{farias02} a method in which, from any function $\Theta:[0,1]^n \rightarrow [0,1]$, it is possible to create a family of $GM$ functions, this method is described by the following proposition.

\begin{prop}[Theorem 2 of \cite{farias02}]\label{def:TeoremaConstrucao}
	Let $\Theta:[0,1]^n \rightarrow [0,1]$ be a function. Then 	{\small
		\begin{equation}
		H_\Theta(x_1, \cdots, x_n) = \left\{
		\begin{array}{ll}
		\displaystyle x_1, & \hbox{if } x_1 = \cdots = x_n\\
		\displaystyle \frac{1}{n-1} \sum_{i=1}^{n}\left(x_i - \frac{x_i |x_i - \Theta(x_1, \cdots, x_n)|}{\displaystyle \sum_{j=1}^{n}|x_j - \Theta(x_1, \cdots, x_n)|} \right),  & \hbox{otherwise}
		\end{array}
		\right.
		\end{equation}} is a GM function.
\end{prop}

Thus, Farias \textit{et al.} have established a family of  generalized mixture functions, such as $H_{Med}, H_{Arith}, H_{Max}, H_{Min}$ and $H_{Mode}$ which can be seen in \cite{farias02} and \cite{farias04}. In this paper, we use the $H_{Med}, H_{Arith}$ and $H_{Max}$ functions as a combination method to combine the output of ensembles.

\begin{eqnarray}
	\small
	H_{Med}(x_1, \cdots, x_n) & = & \left\{
	\begin{array}{ll}
		\displaystyle x_1, & \hbox{if } x_1 = \cdots = x_n\\
		\displaystyle \frac{1}{n-1} \sum_{i=1}^{n}\left(x_i - \frac{x_i |x_i - Med(x_1, \cdots, x_n)|}{\displaystyle \sum_{j=1}^{n}|x_j - Med(x_1, \cdots, x_n)|} \right),  & \hbox{ otherwise.}
	\end{array}
	\right.\\
	H_{Arith}(x_1, \cdots, x_n) & = & \left\{
	\begin{array}{ll}
		\displaystyle x_1, & \hbox{if } x_1 = \cdots = x_n\\
		\displaystyle \frac{1}{n-1} \sum_{i=1}^{n}\left(x_i - \frac{x_i |x_i - Arith(x_1, \cdots, x_n)|}{\displaystyle \sum_{j=1}^{n}|x_j - Arith(x_1, \cdots, x_n)|} \right),  & \hbox{ otherwise.}
	\end{array}
	\right.\\
	H_{Max}(x_1, \cdots, x_n) & = & \left\{
	\begin{array}{ll}
		\displaystyle x_1, & \hbox{if } x_1 = \cdots = x_n\\
		\displaystyle \frac{1}{n-1} \sum_{i=1}^{n}\left(x_i - \frac{x_i |x_i - Max(x_1, \cdots, x_n)|}{\displaystyle \sum_{j=1}^{n}|x_j - Max(x_1, \cdots, x_n)|} \right),  & \hbox{ otherwise.}
	\end{array}
	\right.	
\end{eqnarray}

\begin{rmk}
		The functions $H_{Med}, H_{Arith}, H_{Max}, H_{Min}$ and $H_{Mode}$ also satisfy the properties described in the propositions 2.3 and 2.4. 
\end{rmk}

For a more comprehensive study about the generalized mixture functions, readers are strongly suggested to refer to \cite{farias01,farias02,farias03,farias04,farias05,pere01,pere02}. 

\section{Classifier Ensembles}
\begin{figure}[h]
	\centering
	\includegraphics[width=0.45\linewidth]{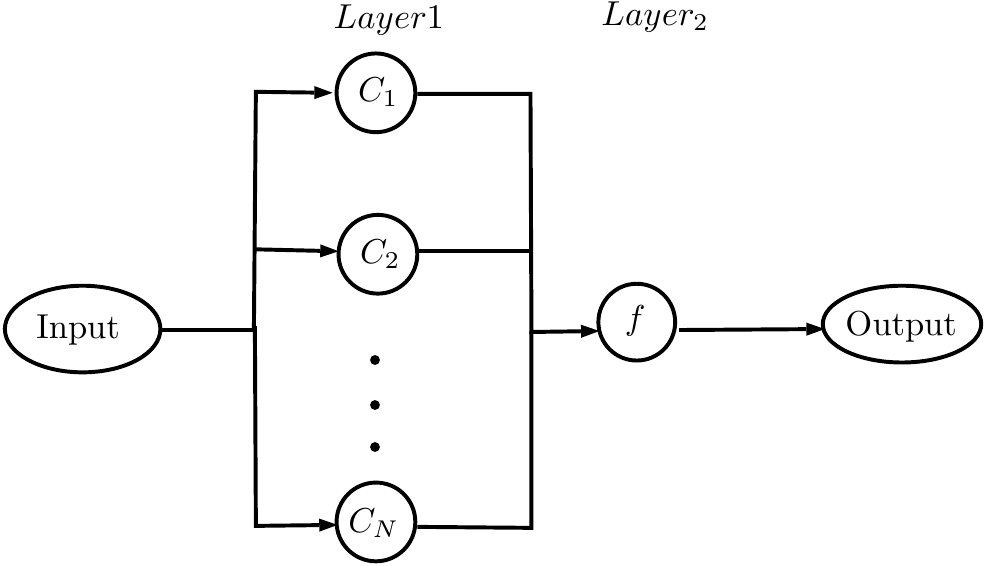}
	\caption{Illustration of the general architecture of an ensemble system.}
	\label{Ima1}
\end{figure}

As mentioned previously, classifier ensembles can be seen as a group of individual classifiers (or experts) that are combined in order to solve a classification problem in a more efficient way. In general, an ensemble can be considered as a two-layer pattern recognition structure \cite{kun04}, as illustrated by figure \ref{Ima1}. The first layer corresponds to the individual classifiers, the members of an ensemble. In this layer, all individual classifiers receive the input pattern and provide an output. In this part, the individual classifiers can be built using the same training dataset, in a redundant approach \cite{kun04}. However, the individual classifiers can be built using a different training set, leading to a multi-view learning approach, in which distinct feature sets are assigned to each classifier but all of them perform the same task \cite{CHAO201601,CHAO201602,CHAO201603}, or to a multi-task learning approach \cite{Sun2013}, in which different features and tasks are assigned to each classifier. In this study, the redundant approach is used. However, this work could be easily extended to a multi-view or multi-task approach.

The second layer of an ensemble represents the combination method, which receives the outputs of all individual classifiers and provides the final output of an ensemble. The main focus of this paper is in the second level, the combination module.

For the combination module, there are three main strategies to combine the output of the individual classifiers,  which are:

\begin{enumerate}
	\item Fusion methods: For this strategy, the combination module takes into consideration all the individual classifiers when defining the final answer of an ensemble. In other words, the combination module uses a mathematical tool to combine the outputs of the first layer, and thus to generate the output of the classifier ensemble \cite{ale01,kiter01,tax01}; 
	\item Selection methods: For this strategy, one individual classifier is selected and is used as a guide to provide the output of the classifier ensemble \cite{kun04}. Therefore, the final output of an ensemble can be defined as the output of the selected individual classifier; 
	\item Hybrid methods: For this strategy, both strategies cited above are used. Usually, for hybrid methods, its first choice is selection but the selection strategy is picked if and only if the selected classifier is considered efficient to correctly classify the corresponding input pattern. If not, a fusion method is then selected. 
\end{enumerate}

In this paper, we apply a generalized mixture functions as a combination method for a classifier ensembles, applying the fusion strategy described above. Therefore, a more detailed description of this strategy is done in this section.

Suppose we have an ensemble system with $N$ members and we apply this system to the classification of an input instance $x$, assigning a class label to this instance, among $L$ different class labels. In this scenario, we can describe the behavior of a combination (fusion) method as follows. 

In the first step of processing, an input instance $x$ is presented to the all individual classifiers of this ensemble. Then, each classifier $C_i$ generates an output for this instance, represented by a vector $O_i$ of dimension $L$, where the $j$-\textit{th} position of this vector is written as $O_i^j$. This position represents the posterior probability assigned by classifier $C_i$ to input pattern $x$ and it defines the degree in which this instance belongs to class $j$, such that $1 \leq i \leq N$ and $1 \leq j \leq L$. 

\begin{figure}[h]
	\centering
	\includegraphics[width=0.9\linewidth]{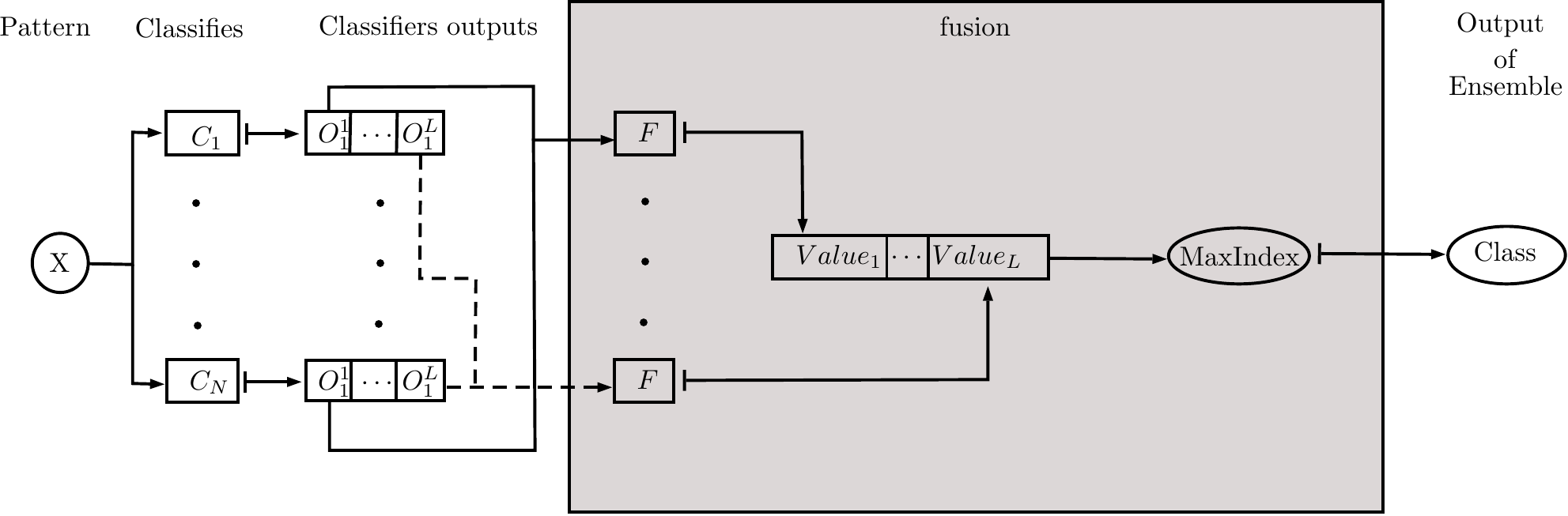}
	\caption{Operating diagram of an ensemble system, adapted from  \cite{ale01}}
	\label{Ima2}
\end{figure}

In the second step the classification process, the combination module $F:[0,1]^N \rightarrow [0,1]$ is applied $L$ times to combine the $i$-th position of all vectors. Then, we have $F(O_1^j, \cdots, O_N^j)$ for each $i = 1,2,\cdots, N$. In other words, the combination method generates the following vector $(F(O_1^1, \cdots, O_N^1), \cdots, F(O_1^L, \cdots, O_N^L))$. Finally in the third step, the output of an ensemble is a class $y$ such that we have the following equation

\begin{equation}
F(O_1^y, \cdots, O_N^y) = max(F(O_1^1, \cdots, O_N^1), \cdots, F(O_1^L, \cdots, O_N^L)) 
\end{equation}
where $1 \leq y \leq L$ and the output of the ensemble is the class assigned to the input example. 
In this case,  the class with the highest value obtained by the combination function is selected. In general, if $max(F(O_1^1, \cdots, O_N^1), \cdots, F(O_1^L, \cdots, O_N^L)) = F(O_1^i, \cdots, O_N^i)$ but $F(O_1^i, \cdots, O_N^i) = F(O_1^j, \cdots, O_N^j)$ with $i \neq j$, the class is chosen randomly between $i$ and $j$. This whole process can be better understood in Algorithm \ref{alg:1} or by the diagram in Figure \ref{Ima2}.

\begin{algorithm}
	\caption{Fusion of classifiers using a function $F$}\label{alg:1}
	\begin{algorithmic}[1]
		\Procedure{Classify-Instance-By-Fusion}{$X$}\Comment{ $X$ is the pattern to be classified.}
		\State $O_1, \cdots, O_N \gets (0,\cdots,0)$\Comment{vectors of size $L$, set initially to 0, $N$ is a number of classifiers.}
		\For{$i=1$ to $N$}
		\State $O_i \gets$ Classify-Instance$_i$(X)\Comment{$O_i$ gets the distribution of posterior probabilities of $i$-th classifier.}
		\EndFor
		\State $Value \gets (0,\cdots,0)$\Comment{vector of $L$ dimensions set initially to 0, that represents the output combination.}
		\For{$j=1$ to $L$}
		\State $Value_i \gets F(O_1^j, \cdots, O_N^j)$\Comment{$O_i^j$ is the $j$-\textit{th} element of the vector $O_i$.}
		\EndFor
		\State \textbf{return} $MaxIndex(Value)$\Comment{$MaxIndex$ returns the index of the highest value in the vector, that is class of output.}
		\EndProcedure
	\end{algorithmic}
\end{algorithm}

\begin{exem}\label{exe:1}
		Suppose that we have an ensemble with 3 classifiers ($C_1, C_2$ and $C_3$), that is applied to a 2-class classification problem. Given an instance $X$, suppose that  $C_1, C_2$ and $C_3$ provide the following output vectors $O_1 = (0.45, 0.55), O_2 = (0.3, 0.7)$ and $O_3 = (0.5, 0.5)$. Then the combination process using the $min$ operator can be described as follows: 
		\begin{itemize}
			\item The combination to the class 1,
			\begin{eqnarray}
				F(O_1^1, O_2^1, O_3^1) & = & min(O_1^1, O_2^1, O_3^1) \nonumber \\
				& = & min(0.45, 0.3, 0.5) \nonumber \\
				& = & 0.3
			\end{eqnarray}
			\item The combination to the class 2,
			\begin{eqnarray}
				F(O_1^2, O_2^2, O_3^2) & = & min(O_1^2, O_2^2, O_3^2) \nonumber \\
				& = & min(0.55, 0.7, 0.5) \nonumber \\
				& = & 0.5
			\end{eqnarray}
			then $value = (0.3, 0.5)$, so $MaxIndex(value) = 2$. Therefore, the input pattern is classified according to the ensemble as belonging to class 2.
		\end{itemize}
\end{exem}


\section{Generalized mixture (GM)-based Combination method}\label{Aplicacao}

As mentioned previously, along with the outputs of the individual classifiers, the combination methods can also receive a set of weights as input, also referred to as weighted-based combinations \cite{kun04}. These weights can represent the confidence of the classifiers in the classification process of a combination method. Basically all combination methods can use weights in their functioning. In order to apply a weighted-based combination method, it is necessary to define this set of weights. Usually, it is defined during the training phase of an ensemble system and they are used throughout the testing phase. In other words, after the definition of the set of weights, they are kept constant during the whole testing phase of an ensemble (static weighting process). However, suppose that an individual classifier received the lowest weight of all individual classifiers for a specific class. In this case, the decision made for this classifier in this class has low chance to be considered by an ensemble, for all testing patterns of this class, since the opinion of the classifier has a small weight.

On the other hand, one of the main advantages of the generalized mixture (GM) functions is that the set of weights can be dynamically defined for each testing pattern. Therefore, the use of GM functions as a combination method eliminates the main problem of the static weighting process, which is the need to define a priori the weights of each individual classifier. 

As already presented in Proposition \ref{def:TeoremaConstrucao}, the GM functions are constructed using as basis the function $\Theta$, that we call a referential point, for each class. In order to use the GM functions as a combination method in an ensemble (GM-based combination), we assumed that the referential point is an approximation of a consensus among the opinions of all classifiers in an ensemble system. In other words, the referential point is represented by the output of one individual classifier that represents a consensus among the opinions of all classifiers. The selection of the referential point is made for each testing pattern. Once the referential point is defined, the GM functions can be easily adapted to be used as a combination method.

The main aim of the GM function is to define the set of weights based on a defined referential point. Therefore, the weights of an individual classifier for a class is defined, based on the distance of the class output of this classifier from the referential point (selected classifier), in an inversely proportional way. In other words, the classifiers whose outputs are distant of the referential point have low weights, while outputs close to the referential point have high weights. The definition of the referential point can be done by any GM function $H_\Theta$. For $H_{Med}$ (Eq.(7)), for instance, the referential point is the median value of all outputs of a class. On the other hand, in $H_{Arith}$ (Eq.(8)) and $H_{Max}$ (Eq.(9)), the referential points are the average and maximum output values, respectively.

\begin{figure}[!h]
	\centering
	\includegraphics[width=0.7\linewidth]{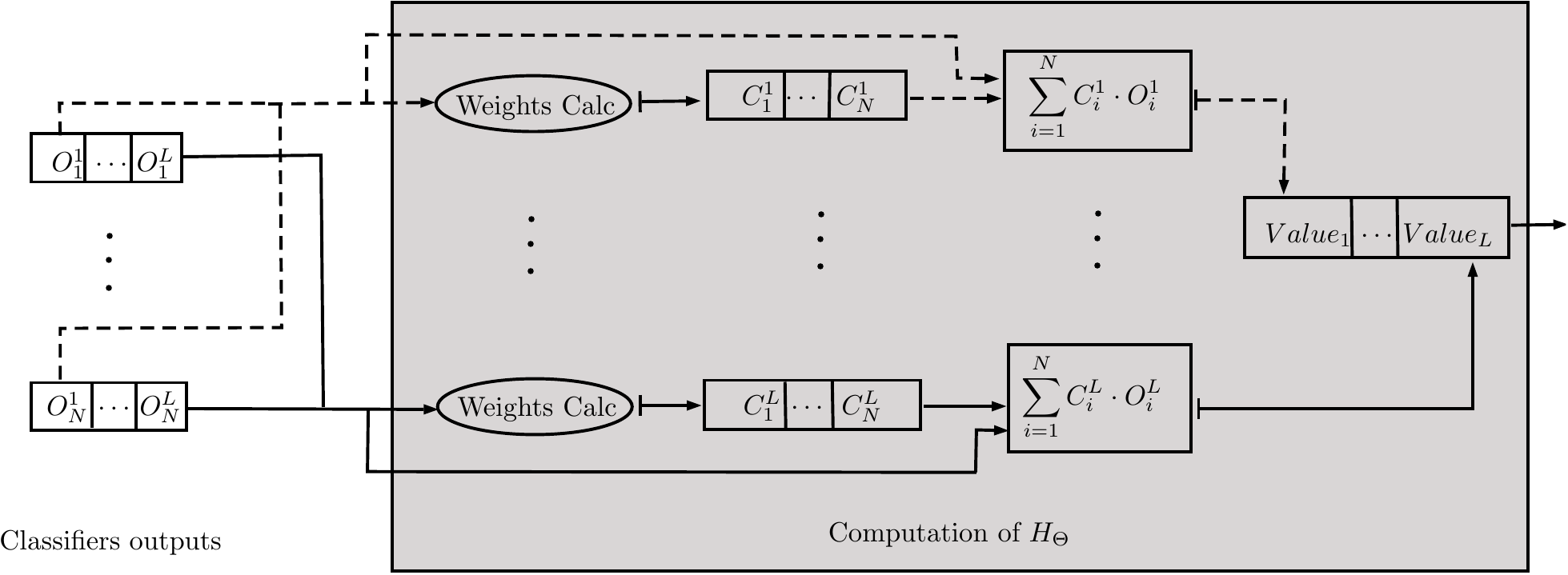}
	\caption{Operating diagram of the GM-based combination method $H_\Theta$.}
	\label{FiguraNova}
\end{figure}

Figure \ref{FiguraNova} illustrates the general functioning of the GM-based combination method. As it can be observed in Figure \ref{FiguraNova} that given a testing pattern, for each class of a problem, the weights are calculated (Weight Calc module), based on a referential point. Then, these weights are used, along with the classifier outputs, to define the final output for a testing pattern ($value_i$). 

Algorithm \ref{alg:2} presents the GM-based combination method. Given a testing pattern,  the weights of each classifier is updated $L$ times (see loop starting at line 7), where $L$ is the number of classes (labels) of the problem.  Then, these weights are used, along with the classifier outputs, to provide the final ensemble output.

\begin{algorithm}
	\caption{Combination method using a function $H_\Theta$.}\label{alg:2}
	\begin{algorithmic}[1]
		\Procedure{Classify-Instance-By-Fusion-$H_\Theta$}{$X$}\Comment{$X$ is the input pattern to be classified.}
		\State $O_1, \cdots, O_N \gets (0,\cdots,0)$\Comment{vectors of size $L$, set initially to 0, $N$ is a number of classifiers.}
		\For{$i=1$ to $N$}
		\State $O_i \gets$ Classify-Instance$_i$(X)\Comment{$O_i$ gets the distribution of posterior probabilities of $i$-th classifier.}
		\EndFor
		\State $Value \gets (0,\cdots,0)$\Comment{vector of $L$ dimensions set initially to 0, that represents the output fusion.}
		\State $weight \gets (0,\cdots,0)$\Comment{is an vector of $N$ dimensions initialized with 0's.}
		\For{$i=1$ to $L$}
		\State $w_i \gets WeightsCalc(O_1^i, \cdot, O_N^i)$\Comment{$O_l^j$ is the $j$-\textit{th} element of the vector $O_l$.}
		\State $Value_i \gets \displaystyle\sum_{j=1}^{N} (O_1^j \cdot w_j)$
		\EndFor
		\State \textbf{return} $MaxIndex(Value)$\Comment{$MaxIndex$ returns the index of the highest value in the vector, that is class of output.}
		\EndProcedure
	\end{algorithmic}
\end{algorithm}

The use of a generalized mixture function $H_\Theta$ as a combination method, can be described by two steps. First, the weight of the each classifier are calculated, using all outputs for a specific class and, secondly, in the weighted average computation (Eq.(\ref{eq:2})). These two steps are respectively performed at lines 8 and 9 of Algorithm \ref{alg:2}, and these steps are the main difference from Algorithm \ref{alg:2} to algorithm \ref{alg:1}. The algorithm for the weight calculation function (line 9 in Algorithm \ref{alg:2}) is presented by Algorithm \ref{alg:3}.

\begin{algorithm}
	\caption{Weight calculation of a GM-based combination method.}\label{alg:3}
	\begin{algorithmic}[1]
		\Procedure{WeightsCalc}{$o_1,\cdots, o_N$}\Comment{$o_i \in [0,1]$ for all $i \leq N$.}
		\State $W \gets (0,\cdots,0)$\Comment{is a vector of $N$ dimensions initialized with 0's.}
		\State $\alpha \gets \Theta(o_1,\cdots, o_N)$\Comment{Computation of the function $\Theta$ to select the referential point.}
		\State $d \gets \displaystyle\sum_{i=1}^{N}|o_i - \alpha|$
		\While{$i \leq N$}\Comment{Computation of the weights of each classifier.}
		\If{$d > 0$}
		\State $w_i \gets \displaystyle\frac{1}{N-1}\PC{1 - \PC{\frac{|o_i-\alpha|}{d}}}$ 
		\Else
		\State $w_i \gets \displaystyle\frac{1}{N}$ 
		\EndIf
		\EndWhile
		\State \textbf{return} $W$\Comment{The weights vector.}
		\EndProcedure
	\end{algorithmic}
\end{algorithm}

For a better understanding of the behavior of the $H_\Theta$ function as a combination method, in the Example \ref{exe2}, a simple application of the functioning of a GM-based combination method is shown.

\begin{exem}\label{exe2}
	Suppose that we have an ensemble with 3 classifiers ($C_1, C_2$ and $C_3$). Additionally, consider that this ensemble is applied to a 2-class classification problem. Given a testing instance, $x$,  classifiers  $C_1, C_2$ and $C_3$ respectively generate three output vectors $O_1 = (0.9, 0.1), O_2 = (0.3, 0.7)$ and $O_3 = (0.5, 0.5)$. Then the combination function $H_{Arith}$ works as follows.
	\begin{itemize}
		\item First, the weights of class $1$ are defined (line 3 of Algorithm \ref{alg:3}). For $H_{Arith}$, the referential point of this class is defined as follows.
		\begin{equation}
		\displaystyle\alpha = \frac{1}{3}\sum_{i=1}^{3}O_i^1 = \frac{0.9+0.3+0.5}{3} = 0.5667
		\end{equation}
		
		The sum of distances (line 4 of Algorithm \ref{alg:3}) is defined as follows.
		\begin{equation}
		\displaystyle d = \sum_{i=1}^{3}|O_i^1 - \alpha| = 0.3333 + 0.2667 + 0.0667 = 0.6667
		\end{equation}
		
		then, the weights of each classifier in relation to class 1 (lines 6-9 of Algorithm \ref{alg:3}) are defined as, 
		\begin{eqnarray}
			w_{1} & = & \frac{1}{2}\PC{1 - \PC{\frac{|O_1^1-\alpha|}{d}}} = \frac{1}{2}\PC{1 - \PC{\frac{|0.9-0.5667|}{0.6667}}} = \frac{1}{2}\PC{1 - \PC{\frac{0.3333}{0.6667}}} \approx 0.25 \nonumber \\
			w_{2} & = & \frac{1}{2}\PC{1 - \PC{\frac{|O_2^1-\alpha|}{d}}} = \frac{1}{2}\PC{1 - \PC{\frac{|0.3-0.5667|}{0.6667}}} = \frac{1}{2}\PC{1 - \PC{\frac{0.2667}{0.6667}}} \approx 0.30 \nonumber \\
			w_{3} & = & \frac{1}{2}\PC{1 - \PC{\frac{|O_3^1-\alpha|}{d}}} = \frac{1}{2}\PC{1 - \PC{\frac{|0.5-0.5667|}{0.6667}}}  = \frac{1}{2}\PC{1 - \PC{\frac{0.0667}{0.6667}}} \approx 0.45 \nonumber
		\end{eqnarray}
		and the ensemble output for  class 1 is defined as follows.
		\begin{eqnarray}
			Value_1 & = & \sum_{i=1}^{3} O_i^1 \cdot w_{i} \nonumber \\
			& = & (0.9 \cdot 0.25) + (0.3 \cdot 0.30) + (0.5 \cdot 0.45) \nonumber \\
			& = & 0.54
		\end{eqnarray}
		\item For class $2$, the referential point is defined as follows.
		\begin{equation}
			\displaystyle\alpha' = \frac{1}{3}\sum_{i=1}^{3}O_i^2 = \frac{0.1+0.7+0.5}{3} = 0.433333
		\end{equation}
		
		and, 
		
		\begin{equation}
			\displaystyle d' = \sum_{i=1}^{3}|O_i^2 - \alpha| = 0.333333 + 0.266667 + 0.066667 = 0.666667
		\end{equation}
		
		the weights of each classifier in relation to class 2 are defined as follows. 
		\begin{eqnarray}
			w_{1} & = & \frac{1}{2}\PC{1 - \PC{\frac{|O_1^2-\alpha'|}{d'}}} = \frac{1}{2}\PC{1 - \PC{\frac{|0.1-0.433333|}{0.666667}}} = \frac{1}{2}\PC{1 - \PC{\frac{0.333333}{0.666667}}} \approx 0.25 \nonumber\\
			w_{2} & = & \frac{1}{2}\PC{1 - \PC{\frac{|O_2^2-\alpha'|}{d'}}} = \frac{1}{2}\PC{1 - \PC{\frac{|0.7-0.433333|}{0.666667}}} = \frac{1}{2}\PC{1 - \PC{\frac{0.266667}{0.666667}}} \approx 0.30 \nonumber \\
			w_{3} & = & \frac{1}{2}\PC{1 - \PC{\frac{|O_3^2-\alpha'|}{d'}}} = \frac{1}{2}\PC{1 - \PC{\frac{|0.5-0.433333|}{0.666667}}}  = \frac{1}{2}\PC{1 - \PC{\frac{0.066667}{0.666667}}} \approx 0.45 \nonumber
		\end{eqnarray}
		and the ensemble output for  class $2$ is defined as,
		\begin{eqnarray}
			Value_2 & = & \sum_{i=1}^{3} O_i^2 \cdot w_{i} \nonumber \\
			& = & (0.1 \cdot 0.25) + (0.7 \cdot 0.30) + (0.5 \cdot 0.45) \nonumber \\
			& = & 0.46
		\end{eqnarray}
	\end{itemize}
	Therefore, as $Value = (0.54, 0.46)$, then $MaxIndex(Value) = 1$. In this sense, this testing  pattern is classified as belonging to class 1, according to the ensemble system using $H_{Arith}$ as a GM-based combination method.
\end{exem}

\section{Experimental Methodology}

In order to evaluate the feasibility of the proposed approach as a combination module of an ensemble system, an empirical analysis is conducted. In this analysis, the obtained ensembles are applied to 25 classification data sets, extracted from UCI \cite{uci01} and other repositories. Table \ref{tab2} presents a description of the used data sets, describing the number of instances, classes and attributes for each data set.

\begin{table}[h]
	\footnotesize
	\centering
	\begin{tabular}{c|c|c|c}
		\hline
		Data bases & Number of instances & Number of Attributes & Classes \\ \hline
		annel.ORIG & 898 & 39 & 6 \\ 
		breast-cancer & 286 & 10 & 2 \\
		cars & 1728 & 7 & 4 \\ 
		german-credit & 1000 & 21 & 2 \\
		glass & 214 & 10 & 7 \\ 
		horse-colic.ORIG & 368 & 28 & 2 \\
		hypothyroid & 3772 & 30 & 4 \\
		ionosphere & 351 & 35 & 2 \\ 
		iris & 150 & 5 & 3 \\ 
		kr-vs-kp & 3196 & 37 & 2 \\
		mfeat-Fourier & 2000 & 77 & 10 \\
		nursery & 12960 & 9 & 5 \\
		optdigits & 5620 & 65 & 10 \\
		pima-diabetes & 768 & 9 & 2 \\
		segment & 2310 & 20 & 7 \\
		sick & 3772 & 30 & 2 \\ 
		soybean & 683 & 36 & 19 \\
		spambase & 4601 & 58 & 2 \\
		splice & 3190 & 62 & 3 \\
		tic-tae-toe & 958 & 10 & 2 \\
		vehicle-silhoettes & 946 & 19 & 4 \\
		vote & 435 & 17 & 2 \\ 
		waveform & 5000 & 41 & 3 \\
		yeast & 1484 & 9 & 10 \\
		zoo & 101 & 18 & 7 \\ \hline
	\end{tabular}
	\caption{Database list used}
	\label{tab2}
\end{table}

In this analysis, seven practical scenarios are used, varying the number of individual classifiers. Therefore, the seven size scenarios refer to as $N =$ \{$5,7,10,15,20,30,40,50$\}, where $N$ is the number of individual classifiers of an ensemble. In other words, the ensemble size varies from 5 to 50 individual classifiers. In fact, in an initial analysis, we applied the proposed approach for ensembles with more individual classifier (larger ensembles). However, as the behavior of the classifier ensembles were very similar to the ones delivered by ensembles with 20 individual classifiers, we decided to use only these ensemble sizes. 

In this analysis, we use heterogeneous ensembles composed of five classification algorithms used as individual classifiers,which are  {\it k}-NN ({\it k} nearest neighbor), Decision Tree, MLP neural network, Naive Bayes as well as Support Vector Machine (SVM). These algorithms are used since they are simple and efficient algorithms and have been widely used in classifier ensembles. All these algorithms are exported from the Weka machine learning package \cite{weka01} and they are applied with the default parameter setting.  For ensemble size $7$, they are composed of 2 {\it k}-NN and 2 decision trees and 1 MLP, SVM and Naive Bayes. For the other ensemble sizes,  we have the same proportion of each classification algorithm. As we are applying a re-sampling procedure similar to Bagging, there was no need to change the parameter setting of the individual classifier for ensembles with more than 5 members.

For the combination module, the proposed generalized function uses the following functions $H_{Max}$, $ H_{Arith}$ and $H_{Med}$, defined in Section \ref{sec:prop}. For comparison purposes, the classifier ensembles with the proposed approach are compared to systems composed of the following static fusion methods: Maximum (Max), Arithmetic mean (Arith), Product (Prod), Majority vote (Vote). In addition, a 10-fold cross-validation is used in which 9 folds are used for training and one for test \cite{koha01}. Additionally, each data set is performed 10 times, leading to a total of 100 runs for each method in each data set.

In order to validate the performance of the classifier ensembles in a more statistically significant way, we apply the Friedman test and Nemenyi post-hoc test, since these non-parametric tests are suitable to compare the performance of different learning algorithms (for a thorough discussion on these statistical tests, please refer to~\cite{Demsar2006}). Both tests are applied considering the results of all ensemble for all 100 runs.


\section{An Analysis of the Experimental Results}

In this paper, the performance of the classifier ensembles using the generalized mixture functions as a combination method is evaluated. In order to do this, the average accuracy of a classifier ensemble is used as the evaluation metric, for all 25 different data sets. In other words, for simplicity reasons, the accuracy of a specific classifier ensemble is averaged over all 25 data sets.

In this section, we divide the result analysis into three parts. The first two parts analyzes the performance of the proposed methods, assessing the impact of the ensemble sizes in the performance of the classifier ensembles and a comparative analysis of all classifier ensembles. In the last part, the proposed method is compared to some of the state-of-the-art ensemble generation methods.

\subsection{Ensemble Size}

In this first analysis, we evaluate the overall average accuracy of the classifier ensembles, when increasing the number of members in an ensemble. Table \ref{tab1} presents the overall performance of each ensemble when varying the number of members. In this table, the overall accuracy and standard deviation of all combination methods are presented to all seven analyzed ensemble sizes. The last line of this table presents the average accuracy, for each combination method.  

\begin{table}[!h]
	\scriptsize
	\centering
	\begin{tabular}{c|c|c|c|c|c|c|c|c}
		\hline
		Size & $H_{Max}$ & $H_{Arith}$ & $H_{Med}$ & $Max$ & $Arith$ & $Prod$ & $Vote$ & $Best_{Comb}$\\ \hline
		5     & 0.882 $\pm$ 0.028&0.882 $\pm$ 0.029&0.874 $\pm$ 0.029&0.857 $\pm$ 0.030&{\bf 0.915 $\pm$ 0.026}&0.882 $\pm$ 0.028&0.873 $\pm$ 0.032 & $Arith$ \\ \hline
		7     & {\bf 0.877 $\pm$ 0.030}&{\bf 0.877 $\pm$ 0.030}&0.875 $\pm$ 0.030&0.857 $\pm$ 0.031&0.871 $\pm$ 0.047&0.863 $\pm$ 0.030&0.871 $\pm$ 0.074 & $H_{Max}, H_{Arith}$ \\ \hline
        10      & {\bf 0.882 $\pm$ 0.029}&{\bf 0.882 $\pm$ 0.029}&{\bf 0.882 $\pm$ 0.029}&0.835 $\pm$ 0.025&0.874 $\pm$ 0.034&0.785 $\pm$ 0.046&0.881 $\pm$ 0.029 &  $H_{Max}, H_{Arith},H_{Med}$ \\ \hline
        15      & {\bf 0.870 $\pm$ 0.132}&0.868 $\pm$ 0.136&0.868 $\pm$ 0.133&0.832 $\pm$ 0.032&0.848 $\pm$ 0.044&0.778 $\pm$ 0.053&0.860 $\pm$ 0.037 & $H_{Max}$\\ \hline
        20   &   0.889 $\pm$0.090&0.890$\pm$0.093&{\bf 0.893$\pm$0.089}&0.827 $\pm$ 0.092&0.851 $\pm$ 0.098&0.779 $\pm$ 0.106&0.849 $\pm$ 0.094 & $H_{Med}$\\ \hline	
        
        30 & 0.859 $\pm$ 0.024 & 0.856 $\pm$ 0.024 & 0.856 $\pm$ 0.024 & 0.853 $\pm$ 0.114 & {\bf 0.875 $\pm$ 0.103} & 0.860 $\pm$ 0.113 &0.854 $\pm$ 0.110 & $Arith$ \\ \hline
        
        40  & 0.890 $\pm$ 0.090 & 0.891 $\pm$ 0.093 & {\bf 0.895 $\pm$ 0.089} & 0.855 $\pm$ 0.114 & 0.869 $\pm$ 0.106 & $ 0.859 \pm 0.114 $ & $ 0.855 \pm 0.110 $ & $H_{Med}$\\ \hline
        50      &  0.883 $\pm$ 0.092  & {\bf 0.901 $\pm$ 0.082} & 0.884 $\pm$ 0.094 & 0.854 $\pm$ 0.113 & 0.869 $\pm$ 0.104 & 0.856 $\pm$ 0.114 & 0.853 $\pm$ 0.111 & $H_{Arith}$ \\ \hline    
        Avg & 0.879 $\pm$ 0.064 & {\bf 0.880 $\pm$ 0.065} & 0.878 $\pm$ 0.065 & 0.846 $\pm$ 0.069 & 0.872 $\pm$ 0.07 & 0.833 $\pm$ 0.076 & 0.862 $\pm$ 0.075 & $H_{Arith}$ \\ \hline
       
    \end{tabular}   	
    	\caption{Results (accuracy$\pm$standard deviation) for ensembles with the proposed combination methods and the traditional ones}
    	\label{tab1}
  \end{table}
        

As it can be observed when analysing the different columns of Table \ref{tab1}, the increase of the number of classifiers had different effect in the performance of the different combination methods. For the proposed approaches, there is a increase and stabilization in accuracy (with some small decreases), when increasing the number of classifiers. When moving the ensemble sizes from 5 to 20, there is an increase in accuracy, and, after that, there is a  stabilization in accuracy.  For $H_{Max}$ and  $H_{Med}$, for instance, the ensemble size that provided the highest accuracy level was the one using 40 individual classifiers, while 50 is the ensemble size with the highest accuracy level for $H_{Arith}$. However, from ensemble size 20 to 50, the accuracy levels are very similar, tending to a stabilization in performance of all three proposed combination methods.

The interesting aspect of Table \ref{tab1} is that the performance of the classical combination methods had the opposite pattern of behaviour, decreasing the accuracy level when increasing the number of classifiers. For instance, the decrease in accuracy reached 10 percentage points for the $Prod$ combination method. For all three static combination methods, the ensemble size with the highest accuracy level was 5.


Therefore, based on Table \ref{tab1}, we can state that the increase in the number of classifiers had a positive effect in the performance of the proposed combination method,  $H_{Max}, H_{Arith}$ and $H_{Med}$. On the other hand, the increase in the number of classifiers had a negative effect in the performance of the classical combination methods.

Another important point to note is that the proposed methods have a stable accuracy. We can observe that there are no abrupt changes in the accuracy of the proposed methods, when varying the ensemble size. For the traditional combination methods, the change in performance is much more notable than in the proposed approach. This is an indication that the definition of the ensemble size does not have a strong influence in the performance of classifier ensembles using a GM-based combination method.

\subsection{Combination Methods}

Once we have analysed the performance of the combination methods when varying the ensemble size, in this section, we compare the performance of all combination methods (proposed and static ones), for each ensemble size. In order to do this, the last column of Table \ref{tab1} presents the best combination method for each ensemble size.



When analysing the last column of Table \ref{tab1}, we can observe that the ensemble system with the highest overall accuracy was the one combined by the arithmetic mean, when the classifier ensembles have 5 members. For all other ensemble sizes, the ensemble system with the highest overall accuracy was one of the proposed methods, $H_{Arith}$, $H_{Max}$ or $H_{Med}$. It is important to emphasize that the best accuracy of the proposed approaches was obtained when a large number of classifiers were used, from 7 members. This might be an indication that this type of combination must be used when a high amount of information is available. 

When using 5 members, we believe that the proposed approaches did not have enough information to make an efficient decision and their performance were surpassed by a single combination such as arithmetic mean. However, when using ensembles with 7 or more members, the proposed approaches improved and provided better performance than all classical combination methods. When using ensembles with 20 members, for instance, the difference in performance of $H_{Max}$ and $product$ reached 10 percentage points, which is remarkable in terms of accuracy gain. For ensemble sizes higher than 20, the performance of all combination methods stabilized, maintaining a high difference in accuracy of the proposed methods in relation to the static ones. 

Given that an analysis taking into account only the overall accuracy might not be so conclusive, some statistical tests are applied. The main goal is to analyze the statistical significance of the differences presented by the experimental results of the classifier ensembles presented in this paper. 

In this statistical analysis, we first apply the Friedman test to evaluate the statistical significance of the performance of all combination methods (proposed and static ones). Then, when the Friedman test indicates statistically significant differences among all analyzed combination methods, we apply the Nemenyi post-hoc test, to perform a two-by-two comparison. In this paper, we consider a confidence level of 99\%, $p < 0.01$ to reject the null hypothesis, in both statistical tests. For the Post-hoc test, there are two possible results, which are defined as follows.

\begin{itemize}
	\item $p > 0.01$, do not reject the the null hypothesis [($H_0$)]: In this case, there is no statistically significant difference. In other words, the performance of one ensemble system can not be said to be statistically superior to the other one;
	\item $p < 0.01$, reject the the null hypothesis [($H_0$)] and accept the alternative hypothesis [($H_1$)]: In this case, as we accept the alternative hypothesis, we can state that the performance of both systems are statistically different. Therefore, it can be said that the performance of one ensemble system is statistically superior to the other one.
\end{itemize}

We applied the Friedman test in the accuracy levels of all combination methods and the results detected statistical significant differences in all 25 data sets ($p < 0.01$). Then, we applied the Nemenyi post-hoc test for each two-by-two combinations. In Table \ref{tab6}, the results of the Nemenyi post-hoc test are presented. In this table, as we are interested in the comparison between the proposed approaches and the static traditional ensemble methods, each proposed method is allocated in one line, while each columns represents one traditional method. In addition, each cell of this table is represented in the the $x-y-z$ format, where $x$ represents the number of wins detected by the statistical test in favor of the the line method, $y$ represents the number of draws and $z$ represents the number of statistically significant differences not in favor of the line methods.

\begin{table}[!h]
	\scriptsize
	\centering
	\begin{tabular}{c|c|c|c|c}
		\hline
		& \multicolumn{4}{|c}{Ensembles with 5 members} \\
		\hline
		& $Max$ & $Arith$  & $Prod$ & $Vote$\\  \hline
		$H_{Max}$   & 18 - 6 - 1 & 5 - 6 - 14  & 7 - 6 - 12 & 10 - 6 - 9\\
		$H_{Arith}$ & 18 - 6 - 1 & 4 - 5 - 15  & 5 - 6 - 14 & 8 - 6 - 11\\ 
		$H_{Med}$ 	& 16 - 6 - 3 & 4 - 6 - 16  & 6 - 6 - 13 & 7 - 6 - 12\\ \hline
		& \multicolumn{4}{|c}{Ensembles with 7 members} \\
		\hline
		& $Max$ & $Arith$  & $Prod$ & $Vote$\\  \hline
		$H_{Max}$   & 21 - 1 - 3 & 14 - 1 - 10  & 18 - 1 - 6 & 13 - 1 - 11\\
		$H_{Arith}$ & 21 - 1 - 3 & 13 - 1 - 11  & 18 - 1 - 6 & 18 - 1 - 6\\ 
		$H_{Med}$ 	& 20 - 1 - 4 & 14 - 1 - 10  & 17 - 1 - 7 & 13 - 1 - 11\\ \hline
		& \multicolumn{4}{|c}{Ensembles with 10 members} \\
		\hline
		& $Max$ & $Arith$  & $Prod$ & $Vote$\\  \hline
		$H_{Max}$   & 21 - 2 - 2 & 18 - 2 - 5   & 22 - 2 - 1 & 12 - 2 - 11\\
		$H_{Arith}$ & 20 - 2 - 3 & 16 - 2 - 7   & 21 - 2 - 2 & 12 - 2 - 11\\ 
		$H_{Med}$ 	& 20 - 2 - 3 & 16 - 2- 7    & 21 - 2 - 2 & 13 - 2 - 10\\ \hline
		& \multicolumn{4}{|c}{Ensembles with 15 members} \\
		\hline
		& $Max$ & $Arith$  & $Prod$ & $Vote$\\  \hline
		$H_{Max}$   & 16 - 0 - 9 & 20 - 0 - 5   & 19 - 0 - 6 & 18 - 0 - 7\\
		$H_{Arith}$ & 16 - 0 - 9 & 21 - 0 - 4   & 19 - 0 - 6 & 14 - 0 - 11\\ 
		$H_{Med}$ 	& 16 - 0 - 9 & 20 - 0 - 5   & 19 - 0 - 6 & 14 - 0 - 11\\ \hline
		& \multicolumn{4}{|c}{Ensembles with 20 members} \\
		\hline
		& $Max$ & $Arith$  & $Prod$ & $Vote$\\  \hline
		$H_{Max}$   & 14 - 7 - 4 & 17 - 7 - 1   & 15 - 7 - 3 & 17 - 7 - 1\\
		$H_{Arith}$ & 13 - 8 - 4 & 15 - 8 - 2   & 14 - 8 - 3 & 14 - 8 - 3\\ 
		$H_{Med}$ 	& 13 - 8 - 4 & 15 - 8 - 2   & 14 - 8 - 3 & 15 - 8 - 2\\ \hline 
		& \multicolumn{4}{|c}{Ensembles with 30 members} \\
		\hline
		& 	   $Max$ & $Arith$     & $Product$   & $Vote$ \\  \hline
		$H_{Max}$   & 11 - 5 - 9 & 11 - 5 - 9  & 11 - 5 - 9  & 11 - 5 - 9 \\
		$H_{Arith}$ & 10 - 6 - 9 & 10 - 6 - 9  & 10 - 6 - 9  & 10 - 6 - 9 \\ 
		$H_{Med}$ 	& 10 - 6 - 9 & 10 - 6 - 9  & 10 - 6 - 9  & 10 - 6 - 9\\ \hline
		& \multicolumn{4}{|c}{Ensembles with 40 members} \\
		\hline
		& 	   $Max$ & $Arith$     & $Product$   & $Vote$ \\  \hline
		$H_{Max}$   & 13 - 12 - 0  & 0 - 25 - 0  & 11 - 14 - 0  & 17 - 8 - 0 \\
		$H_{Arith}$ & 14 - 11 - 0  & 0 - 25 - 0  & 13 - 12 - 0  & 18 - 7 - 0 \\ 
		$H_{Med}$ 	& 14 - 11 - 0  & 0 - 25 - 0  & 13 - 12 - 0  & 18 - 7 - 0\\ \hline
		& \multicolumn{4}{|c}{Ensembles with 50 members} \\
		\hline
		& 	   $Max$ & $Arith$     & $Product$   & $Vote$ \\  \hline
		$H_{Max}$   & 22  - 3 - 0  & 0 - 25 - 0  & 1 - 24 - 0   & 4 - 21 - 0  \\
		$H_{Arith}$ & 25 - 0  - 0  & 15 - 10 - 0 & 25 - 0 - 0   & 25 - 0  - 0 \\ 
		$H_{Med}$ 	& 25 - 0  - 0  & 15 - 10 - 0 & 25 - 0 - 0   & 25 - 0  - 0 \\ \hline		
	\end{tabular}
	\vspace{0.1cm}
	\caption{Results of the pairwise statistical test, in the $x-y-z$ format, where $x$ = number of wins of the line methods, $y$ = number of draws and $z$ = number of losses of the line methods. }
	\label{tab6}
\end{table}

From this table, we can detect that the results of the statistical test corroborate with the results of Table \ref{tab1}. When comparing the proposed approaches with the maximum, $Max$, combination method (first column of Table \ref{tab6}), we can observe that the values of $x$ are much higher than $y$ and $z$, showing that the proposed approaches provided better performance than the maximun combination method, from a statistical point of view. This pattern of behavior was observed to all ensemble sizes. 

In relation to the other traditional combination methods ($Arith$, $Prod$ and $Vote$), we can see a slightly similar behavior to maximum method. In other words, the proposed methods achieved better performance than the traditional combination methods, from a statistical point of view, for the majority of ensemble sizes. The only exception is ensembles with $5$ members, in which the traditional methods had better performance, since $y$ is higher than $x$ and $z$. When using ensemble with 50 member, for instance, $H_{Arith}$ and $H_{Med}$ had performance statistically better than $Max$, $Arith$ and $Vote$, for all 25 datasets.

These results showed that the proposed approaches improved the performance of the classifier ensembles and this improvement proved to be statistically significant for the majority of data sets, mainly for large classifier ensembles. 


In summary, based on the empirical analysis of this section, we can conclude that the proposed approaches provide more accurate classifier ensembles, when compared to traditional combination methods. However, the best scenarios for the proposed approaches are those with 20 or more members in classifier ensembles, where a high amount of information is available.

\subsection{Comparative Analysis: Ensemble Generation Methods}

Once the comparison with traditional ensemble techniques was conducted, a comparison with recent and the-state-of-the-art ensemble generation techniques is conducted. In this comparative analysis, we have selected four ensemble generation techniques with different functioning and data processing, which are described as follows. 

\begin{enumerate}
	\item Random Forest (RF): This method has been largely used in the ensemble community. In our implementation, the number of trees varied from 100 to 1000, depending on the used dataset. 
	\item XGBoost: It is a more powerful version of Boosting, originally proposed in \cite{chen02}. XGBoost has presented the-state-of-the-art performance in competitions such as  Kaggle \footnote{https://www.kaggle.com/} e ACM \cite{Vol01}.  In our implementation, the number of gbtrees varied between 20 and 100, depending on the used dataset. 
	\item DES-RRC: This method is a dynamic selection ensemble method based on the randomized reference classifier,
	in order to decide whether or not the base classifier $c_i$ performs significantly better than the random classifier. This method was originally proposed in \cite{wolo01}. However, in this paper, we compare the results provided in the extensive analysis made in \cite{Cruz03}
	\item META-DES (MDES): The META-DES method is a dynamic selection ensemble method based on the assumption that the
	dynamic ensemble selection problem can be considered as a metaproblem \cite{Cruz05}. This meta-problem uses different criteria regarding 	the behavior of a base classifier $c_i$ , in order to decide whether it is competent enough to classify a given test sample. Once again, we compare the results provided in the extensive analysis made in \cite{Cruz03}.
	\item META-DES.Oracle (MDES-O): This method is also a dynamic selection ensemble method and is an extension of MDES, in which a total of 15 sets of meta-features were considered. Following that, a meta-features selection scheme
	using a Binary Particle Swarm Optimization (BPSO) was conducted in order to optimize the performance of the meta-classifier \cite{Cruz04}. Once again, we compare the results provided in the extensive analysis made in \cite{Cruz03}.
	\item P2-SA and P2-WTA: These are two versions of a multi-objective algorithm for selecting members for an ensemble system, proposed in \cite{Fer2016}.
\end{enumerate}

Table \ref{tab11} presents the accuracy levels of all ten ensemble methods. The proposed methods with 40 members are used in this table since it provided the overall best performance. The results of the first two methods (Random Forest and XGBoost) were obtained by an implementation done by us. However, for the remaining five methods, we used the original results, provided in \cite{Cruz03} and \cite{Fer2016}. In these papers, a different experimental methodology was used and this may cause a slight difference in the obtained results. However, we would like to have a general picture of the performance of the proposed methods and we decided to use these methods in this comparative analysis. Additionally, this comparative analysis uses a different group of datasets and some cells of Table \ref{tab11} are empty since the corresponding ensemble methods were not applied to the line datasets.  In this table, the bold numbers represent the ensemble method with the highest accuracy level for a dataset.

\begin{table}[!htb]
	\scriptsize
	\centering
	\begin{sideways}
	\begin{tabular}{|c|c|c|c|c|c|c|c|c|c|c|}
		\hline
		& $H_{Max}$ & $H_{Arith}$ & $H_{Med}$ & RF & XGBoost &DES-RRC&MDES&MDES-O&P2-SA &P2-WTA \\ \hline
		{annel.ORIG}        &88.57$\pm$9.43&88.15$\pm$10.27&88.55$\pm$9.61&89.2$\pm$0.21&\textbf{92.55$\pm$1.67}&--&--&--&--&--\\
		{breast-cancer}      &89.40$\pm$9.07&89.54$\pm$9.08&\textbf{89.59$\pm$8.83}&88.63$\pm$5.31&88.99$\pm$6.19&--&--&--&--&--\\
		{cars}               &88.29$\pm$9.64&88.71$\pm$9.8&89.10$\pm$9.48&89.12$\pm$0.12&\textbf{92.18$\pm$1.35}&--&--&--&--&--\\
		{horse-colic.ORIG}   &88.67$\pm$9.64&88.64$\pm$9.61&\textbf{88.98$\pm$9.6}&85.09$\pm$2.65&86.58$\pm$3.03&--&--&--&63.6&63.4\\
		{german-credit}      &89.33$\pm$8.65&89.10$\pm$8.81&\textbf{89.62$\pm$8.7}&87.68$\pm$6.61&88.22$\pm$6.36&75.83$\pm$2.36&75.55$\pm$1.31&76.58$\pm$1.99&71.7&74.3\\
		{pima-diabetes}      &89.03$\pm$8.77&88.95$\pm$9.17&\textbf{89.39$\pm$8.5}&88.95$\pm$3.25&87.48$\pm$4.71&77.64$\pm$2.73&79.03$\pm$2.24&77.53$\pm$2.24&75.5&76\\
		{glass}              &88.51$\pm$9.32&89.00$\pm$9.27&88.91$\pm$9.41&90.57$\pm$1.32&\textbf{91.44$\pm$1.58}&66.04$\pm$4.23&66.87$\pm$2.99&66.46$\pm$4.22&--&--\\
		{hypothyroid}        &88.85$\pm$9.22&89.23$\pm$9.41&89.27$\pm$8.98&89.97$\pm$2.21&\textbf{92.6$\pm$3.15}&--&--&--&--&--\\
		{segment}            &89.30$\pm$8.95&89.37$\pm$9.28&89.71$\pm$8.97&\textbf{93.08$\pm$5.67}&91.16$\pm$2.98&--&--&--&--&--\\
		{ionosphere}         &88.94$\pm$9.84&89.00$\pm$10.11&89.21$\pm$9.61&90.13$\pm$0.59&\textbf{92.07$\pm$1.23}&88.8$\pm$2.48&89.94$\pm$1.96&89.94$\pm$1.97&--&--\\
		{iris}               &96.05$\pm$8.46&96.92$\pm$8.84&\textbf{97.45$\pm$8.57}&93.03$\pm$1.79&94.18$\pm$2.01&--&--&--&--&--\\
		{kr-vs-kp}           &88.91$\pm$8.57&88.68$\pm$8.46&89.57$\pm$8.21&89.84$\pm$0.02&\textbf{91.82$\pm$1.79}&--&--&--&--&--\\
		{mfeat-Fourier}      &89.58$\pm$8.09&89.61$\pm$8.74&90.23$\pm$7.92&89.51$\pm$0.37&\textbf{91.28$\pm$0.05}&--&--&--&--&--\\
		{nursery}            &89.22$\pm$8.88&89.41$\pm$9.36&\textbf{89.66$\pm$8.87}&88.06$\pm$3.59&87.85$\pm$3.79&--&--&--&--&--\\
		{optdigits}          &89.59$\pm$8.57&89.26$\pm$9.17&\textbf{90.04$\pm$8.69}&89.73$\pm$0.33&89.51$\pm$0.42&--&--&--&--&--\\
		{Liver}&69.22$\pm$8.88&69.41$\pm$9.36&69.66$\pm$8.87&71.06$\pm$3.59&74.85$\pm$3.79&68.01$\pm$4.14&70.08$\pm$3.49&\textbf{72.02$\pm$4.72}&69.3&69.9\\
		{sick}               &89.64$\pm$8.77&89.5$\pm$8.97&89.94$\pm$8.53&89.64$\pm$0.3&\textbf{90.40$\pm$0.22}&--&--&--&--&--\\
		{soybean}            &89.10$\pm$9.65&\textbf{89.54$\pm$9.55}&89.42$\pm$9.33&89.06$\pm$0.24&89.2$\pm$1.84&--&--&--&--&--\\
		{spambase}           &88.68$\pm$9.37&89.28$\pm$8.98&\textbf{89.35$\pm$9.03}&89.18$\pm$0.2&89.25$\pm$0.01&--&--&--&--&--\\
		{Segment}             &97.02$\pm$8.71&96.95$\pm$9.01&96.71$\pm$8.56&\textbf{97.42$\pm$6.07}&97.12$\pm$1.43&96.38$\pm$0.75&96.21$\pm$0.87&96.65$\pm$0.83&93.7&93.9\\
		{tic-tae-toe}        &89.33$\pm$8.86&89.31$\pm$9.41&\textbf{89.99$\pm$8.46}&89.57$\pm$0.7&87.29$\pm$1.47&--&--&--&--&--\\
		{vehicle-silhoettes} &88.95$\pm$9.15&88.81$\pm$9.54&\textbf{89.60$\pm$8.81}&88.85$\pm$2.26&88.66$\pm$2.92&83.34$\pm$1.81&82.75$\pm$1.7&82.87$\pm$1.64&76.5&76.2\\
		{vote}               &89.47$\pm$8.98&89.34$\pm$9.78&89.51$\pm$8.94&\textbf{91.02$\pm$3.23}&90.08$\pm$0.85&--&--&--&--&--\\
		{waveform}           &88.94$\pm$8.90&89.03$\pm$9.57&\textbf{89.39$\pm$8.87}&84.48$\pm$1.31&85.08$\pm$1.17&84.63$\pm$0.48&84.56$\pm$0.36&84.72$\pm$0.49&--&--\\
		{yeast}              &89.32$\pm$8.21&89.28$\pm$8.8&\textbf{89.74$\pm$8.48}&88.4$\pm$2.77&89.5$\pm$12.19&--&--&--&--&--\\ \hline
		Average results      &88.88$\pm$8.98&88.96$\pm$9.29&89.30$\pm$8.87&88.85$\pm$2.19&89.57$\pm$2.6&80.08$\pm$2.37&80.62$\pm$1.86&82.90$\pm$2.26&75.05&75.62\\ \hline
	\end{tabular}
\end{sideways}
	\caption{Comparative results: state-of-the-art methods}
	\label{tab11}
\end{table}
As it can be seen in Table \ref{tab11}, the highest accuracy was obtained by one proposed methods in 13 datasets, being $H_{Med}$ the one with more bold numbers. XGBoost also had an outstanding performance with the highest accuracy in 8 datasets and the overall highest accuracy (last line of Table \ref{tab11}).  
We then applied the Friedman test for the first 5 columns of Table \ref{tab11}, obtaining a $p$-value $=$ $0.021$. In other words, the statistical test did not detect a significant difference in performance of these five ensemble generation techniques. It is important to highlight that $DES-RRC$, $META$, $METAO$, $P2-SA$ and $P2-WTA$ were not included in the statistical test since we only had access to the mean accuracy and standard deviation of these techniques, not allowing the application of the statistical test. 

It is important to emphasize that the results obtained in this paper are promising since XGBoost is a scalable and accurate implementation of gradient boosting machines and it has proven to push the limits of computing power for boosted trees algorithms and has been developed for the sole purpose of model performance and computational speed. In other words, it is a powerful and complex ensemble generation method. On the other hand, the proposed methods are simple and inexpensive combination methods that have reached performance similar performance than XGBoost, from a statistical point of view.

Our final analysis is related to the execution time processing for all ensemble techniques. The execution time was measured in seconds and Table \ref{tab12} presents the execution time of all ensemble methods. In this table, the execution time of all ensemble sizes are presented, including the traditional static combination methods..

\begin{table}[!htb]
	\footnotesize
	\centering

		\begin{tabular}{|c|c|c|c|c|c|c|c|c|c|c|}
			\hline	
			EnsSize&Max&Arith&Prod&Vote&$H_{Arith}$&$H_{Med}$&$H_{Max}$&XGBoost&RF \\ \hline
			5&751.40&611.30&512.30&751.40&664.4&1048.7&1110.1&1102.2&954.2 \\ \hline
			7&955.70&783.50&663.10&955.70&1549.1&1198.1&2367.6&2965.1&1565.1 \\ \hline
			10&1176.20&1268.60&1124.90&1216.20&1585.7&1587.8&2493.9&3120.4&1767.4 \\ \hline
			15&1991.70&1807.90&1617.20&1919.40&2384.1&3948.9&4898.2&3482&2491.6 \\ \hline
			20&2658.60&2391.30&2018.60&2708.30&2615.4&4996.1&5682.4&4624.3&2828.1 \\ \hline
			30&3975.90&3197.40&2967.40&4015.10&5916.2&6077.1&7420.1&8756.6&4805.3 \\ \hline
			40&4901.10&3966.40&3791.50&5207.90&6145.2&11379.7&9075&10221.1&5910.4 \\ \hline
			50&5517.90&5102.20&4922.70&5925.40&6993.1&13101.6&9341.1&12408.3&6534.1 \\ \hline
			Ave&2741.062&2391.07&2202.21&2837.42&3481.65&5417.25&5298.55&5835.00&3357.02\\ \hline
			
		\end{tabular}
	\caption{Comparative results: execution time of the ensemble  methods}
	\label{tab12}
\end{table}

As it can be observed in Table \ref{tab12}, the simple combination methods (Max, Arith, Prod and Vote) achieved the lowest execution times. However, these methods provided the lowest accuracy levels. Then, RF achieved the lowest time, followed closely by $H_{Arith}$. Finally, the three ensemble methods with the highest execution times are, in an ascending order, $H_{Max}$, $H_{Med}$ and XGBoost. 

There are some interesting aspects in Table \ref{tab12} that need to be highlighted. First, this was an expected result since the simple methods provided the lowest time and XGBoost achieved the highest execution time. Second, the $H_{Med}$ execution time is much lower than $H_{Max}$ and $H_{Max}$. $H_{Arith}$ uses the average value as the referential point, while $H_{Max}$ and $H_{Med}$ use the maximum and median, respectively. The calculation of maximum and median needs a sorting procedure, which requires more time than the average calculation procedure. Finally, although random forest (RF) has slight lower execution time than $H_{Arith}$, we can observe that the execution time of $H_{Arith}$ for small ensembles is lower than RF. Only when using 30 and more classifiers that RF had lower execution time. 

In summary, we can conclude that the main disadvantage of a dynamic combination method is the increase in complexity when increasing the ensemble size. Nevertheless, one proposed method, $H_{Arith}$, had a comparable execution time, when compared to RF.

\section{Final Remarks}

In this paper, we studied the problem of combining classifiers in ensemble, proposing a new approach for a combination method based on generalized mixture functions, that are $H_{Max}, H_{Arith}$ and $H_{Med}$. We presented the fundamental notions of the generalized mixture functions as well as its adaptation for a fusion method used in classifier ensembles. In order to evaluate the feasibility of the proposed approach, an empirical analysis was conducted, using classifier ensembles varying from 5 to 50 members and applied to 25 different classification data sets. In this analysis, we compared the use of the proposed approaches to classifier ensembles using traditional combination methods, Maximum (Max), Arithmetic mean (Arith), Product (Prod), Majority vote (Vote). Additionally, the proposed methods were also compared to the state-of-the-art ensemble generation methods.


As a result of this empirical analysis, we concluded that the generalized mixture functions $H_{Max}, H_{Arith}$ and $H_{Med}$ 
can actually be used as a framework for the combination of individual classifiers in the design of accurate classifier ensembles. In addition, we concluded that the proposed combination method has better results when  a larger amount of information (or opinions) is available for fusion. In other words, when we have a relatively large number of members in the classifier ensemble. When compared to the the state-of-the-art ensemble generation methods, the proposed methods had an outstanding performance, outperforming most of these methods that included Random Forest, widely used in for classifier ensembles.

It is important to note that the ensemble used in this paper deal with the single-label classification problems. In other words, the instances of a data set only belong to a unique class of the problem. As future work, we intend to address the multi-label classification  problem and also we are interested in investigate the application of generalized mixture functions in the data clustering problem.

\end{document}